\documentclass{article}
\PassOptionsToPackage{numbers,sort&compress}{natbib}
\usepackage[sglblindworkshop]{neurips_2026}
\workshoptitle{Foundation Models for Temporal Systems}
\usepackage[T1]{fontenc}
\usepackage[utf8]{inputenc}
\usepackage{microtype}
\usepackage{graphicx}
\usepackage{booktabs}
\usepackage{multirow}
\usepackage{amsmath,amssymb}
\usepackage{xcolor}
\usepackage{hyperref}
\usepackage{url}
\usepackage{listings}
\usepackage{enumitem}

\newcommand{\gnum}{G\textsubscript{num}}
\newcommand{\gtext}{G\textsubscript{text}}
\newcommand{\gcode}{G\textsubscript{code}}

\title{Training-Free Continual Learning for\\
Multimodal Forecasting via Prompt Evolution}

\author{%
    Laurent Mombaerts\\
    Amazon Web Services\\
    \And
    Jonathan Taws\\
    Amazon Web Services\\
    \And
    Jacopo Gargano\\
    Amazon Web Services\\
    \And
    Dinesh Gaddam\\
    Amazon Web Services\\
    \And
    Kevin George\\
    Amazon Web Services\\
    \And
    Srijan Tiwari\\
    Amazon Web Services\\
    \And
    Shreyas Rajesh\thanks{Work done while an intern at Amazon Web Services.}\\
    University of California\\
}

\begin{document}
\maketitle

\begin{abstract}

Adapting a forecasting system to a changing world classically means retraining. We ask whether the task of continually learning regimes for time series forecasting can be represented by a piece of text: a guidance block,
  evolved by outcome-driven reflection and used in a frozen LLM forecaster's prompt. On a refreshed multimodal benchmark (weekly search indices, daily currencies, daily commodities, each window paired with web-retrieved
  news), we run GEPA-style prompt evolution under strict contamination control to prevent the data leakage common in time series benchmarks: the forecaster's knowledge cutoff predates every evaluated window. A $\sim$900-word
  evolved guidance carried by Llama-3.3-70B beats Time-Series Foundation Models (TSFMs) on held-out data (GM MASE 0.776 vs.\ 0.898 for TabPFN-TS-3), transfers to more recent models without measurable loss, and costs dollars
  rather than GPU-days to update. We characterize the loop: out-of-loop gains concentrate in the first reflection, in-loop improvements overstate out-of-loop gains $2$--$3\times$ unless an untouched monitor arbitrates, and
  error-only objectives silently suppress the text modality. We provide two extensions. First, compiling the guidance into executable code with a TSFM as a callable tool recovers most of the gain deterministically. Second,
  ablations show the advantage widens at short horizons and that removing the news text does not hurt the guided forecaster, consistent with the
text’s value being distilled into policy during learning. 
We argue that continual prompt evolution is a practical, training-free alternative
  to retraining for multimodal forecasting.
\end{abstract}

\section{Introduction}
\label{sec:intro}

Time series forecasting supports decision-making in numerous high-stakes
domains, and the advent of Time Series Foundation Models
(TSFMs)~\cite{das2024timesfm,ansari2024chronos,hollmann2025tabpfn}
has established a unified, zero-shot forecasting paradigm for numerical
sequences. However, relying solely on structured numerical inputs isolates
these models from the broader real-world narratives that drive structural
breaks and regime shifts; conversely, large language models (LLMs) parse
unstructured context easily but are uneven numerical
forecasters~\cite{gruver2023llmtime,tan2024useful}. Benchmarks that pair
numeric histories with contemporaneous news
context~\cite{timesx2026,cik2024} report that carefully engineered
LLM pipelines can exploit text, but the engineering is manual, static, and
brittle under distribution shift.

This paper treats the problem as one of \emph{continual learning} with the learned state living in prompt rather than weights. When a forecasting system must track a changing world, the standard approach is retraining: fine-tuning the model on recent windows, or re-pretraining the TSFM. We instead freeze every model weight for all the models in use and continually learn a \textbf{guidance text}: a block of up to $\sim$1{,}000 words inputs into the forecaster's prompt, telling it how to approach a window. Learning proceeds through outcome-driven evolutionary loop in the spirit of GEPA~\cite{gepa2025}, a prompt-optimization method in which an LLM \emph{reflects} in natural language on execution traces to propose prompt mutations, and candidates are kept on an instance-level Pareto front rather than winner-takes-all. Concretely: (1) a LLM forecasts training windows with the current guidance, (2) a \emph{reflector} LLM reads the full execution traces (what the news said, what the numbers did, what the model predicted, what actually happened) and rewrites the guidance, (3) an acceptance gate discards rewrites that do not beat their parent on fresh windows, and (4) survivors are scored on a fixed selection set under instance-level Pareto selection. No gradient exists anywhere. Every version of the learned artifact is readable, diffable, and auditable: one can point at any rule and ask \emph{which windows taught you that?}

Our study runs on a refreshed copy of the TimesX multimodal benchmark~\cite{timesx2026} (data through March 2026; \S\ref{sec:setup}), under contamination control by construction: the default forecaster (Llama-3.3-70B, Dec-2023 cutoff) predates every evaluated window. The stronger executor used for transfer and code-compilation experiments (Claude Opus 4.6, Aug-2025 cutoff) is evaluated strictly on windows whose origins postdate its cutoff. The reflector writes only general ex-ante rules outcome-gated on windows it never saw, and final test tiers are spent exactly once.

\textbf{Our contributions are the following:}
\begin{enumerate}[leftmargin=14pt, itemsep=1pt, topsep=2pt]
\item \textbf{A continually learned prompt beats TSFMs on their own
terms} (\S\ref{sec:results}): the evolved guidance \gnum{} makes
Llama-3.3-70B outperform TimesFM-2.5, TabPFN-TS-3, and Chronos-2 on two
held-out tiers, updating for dollars with no weight changes.
\item \textbf{The artifact is portable}
(\S\ref{sec:transfer}): guidance evolved for one forecaster transfers to a stronger one (Claude Opus 4.6) without measurable loss (the transferred arm overall results are statistically indistinguishable from the originally evolved model).
\item \textbf{A measured anatomy of the loop} (\S\ref{sec:learnings}): the depth of prompt evolutions tracks the artifact's capacity, with one reflection capturing most of
  what a relatively small fixed-size text prompt can gain; in-loop improvements overstate
  held-out gains $2$--$3\times$; and accuracy-only objectives teach the forecaster to ignore the news text,
  on two different LLMs.
\item \textbf{Guidance compiles to code} (\S\ref{sec:compiled}): a
coding LLM compiles each guidance into a deterministic function with
TabPFN as a callable tool and an LLM arbiter; the learned composition
beats every naive one and posts the best SearchTrend score.
\item \textbf{Ablations that locate the value} (\S\ref{sec:ablations}):
the advantage widens at short horizons (re-evolving for $h{=}3$ does
not beat truncating); removing the unstructured news text entirely does not hurt the guided forecaster, consistent with its decision-relevant content being distilled into the (predominantly numeric) policy during learning, so inference can skip context retrieval.

\end{enumerate}

\section{Related work}
\label{sec:related}

 \textbf{LLMs and context for time series.} Encoding observations as
  strings enables zero-shot LLM extrapolation~\cite{gruver2023llmtime}.
  Reprogramming approaches adapt frozen LLM
  backbones~\cite{jin2024timellm,zhou2023fpt}, though ablations question
  how much the backbone itself contributes~\cite{tan2024useful}.
  Context-aided benchmarks, CiK~\cite{cik2024} and
  TimesX~\cite{timesx2026}, pair numeric histories with textual context.
  Agentic decompositions such as
  Nexus~\cite{nexus2026} calibrate guidelines from backtests. Our loop
  instead makes the guideline itself the sole, continually evolved
  learning object under explicit selection pressure and overfitting
  monitors. FLAIRR-TS~\cite{flairr2025} refines a forecaster's prompt per
  task at test time, whereas we learn one guidance across all series at
  training time, monitor it out-of-loop, and deploy it frozen.   PostTime~\cite{liu2026posttime} post-trains an LLM (SFT then RLVR) on TimesX
  to revise, preserve, or ignore a TimesFM-2.5 prior; we target the same
  LLM-over-TSFM division of labor (\S\ref{sec:compiled}) without any weight update, learning it
  as an auditable, transferable text artifact.

\textbf{TSFMs.} TimesFM-2.5~\cite{das2024timesfm} operates on univariate numeric histories,
  whereas Chronos-2~\cite{ansari2024chronos} and
  TabPFN-TS-3~\cite{hollmann2025tabpfn} also accept numerical covariates. We
  evaluate all three in their standard numeric-only configurations, as baselines
  and, in the compiled extension, as \emph{tools} the learned guidance decides
  when to trust, blend, or override.


\textbf{Prompt optimization and continual learning.} Our loop
instantiates reflective prompt evolution~\cite{gepa2025}: natural-language reflection over execution traces as the mutation operator, an instance-level Pareto front as the selection operator. Related work includes textual-gradient optimization ~\cite{textgrad2024} and self-refinement~\cite{madaan2023selfrefine}. We contribute a forecasting-specific instantiation with strict temporal integrity, a study of its behaviour, and the compile-to-code extension. Classical continual learning updates weights under
stability--plasticity constraints~\cite{parisi2019continual}; here the
entire learned state is a short text, so updates are cheap, reversible,
versioned, and ``forgetting'' is a visible diff.


\section{Benchmark and evaluation protocol}
\label{sec:setup}

\textbf{Data.} We use a refreshed copy of the TimesX benchmark~\cite{timesx2026} (through 2026-03-31), a leakage-free, automated data generation pipeline of high-quality real-world multimodal time series with textual context: \emph{SearchTrend} (weekly Google search-interest indices, 0--100, $\sim$120 topics), \emph{Currency} (daily USD rates, 10 currencies), and \emph{CommodityPrice} (daily prices, $\sim$60 commodities). Each window provides a 96-step history, a 12-step target, and text available \emph{at the forecast origin}: a background description, $\sim$8--10 dated news events as prose, sibling-series statistics, and holidays in the prediction window. Forecasts are scored by MASE (error scaled by the series' typical seasonal-step change, so SeasonalNaive $\approx$1), aggregated as geometric means (GM); contrasts are paired per-window deltas with standard errors. 

\textbf{Contamination control and tiers.} Evaluating LLMs on windows
preceding their cutoff measures capability \emph{plus} memory: LLM forecast error rises markedly once windows postdate the model's cutoff \cite{timesx2026}. Our default forecaster is Llama-3.3-70B
(Dec-2023 cutoff, clean on every evaluated window); the reflector
(Claude Opus 5) is not required to have a cutoff preceding the data, but only writes
only general, ex-ante rules (no variable names, no memorized outcomes)
that are outcome-gated on windows it never saw. Reflector leakage is enforced against three ways: (1) binding prompt constraints (``ex-ante only, no variable names from the evidence, no memorized post-origin outcomes''), (2) structurally, guidance is plain text applied to windows disjoint from those reflected on, so a memorized specific has no key to fire on without naming a series or date, and (3) auditability, every artifact is human-readable, was read, and contains no post-origin dates, values, or events. Cross-model experiments
with Claude Opus 4.6 (Aug-2025 cutoff) use only origins after
September 2025. Optimization consumes data at three rates: a
\emph{selection} tier consulted every iteration (optimistic by
construction); an \emph{untouched monitor} never used for decisions,
whose gap to selection measures overfitting; and \emph{final} tiers
set postdating every model's cutoff touched exactly once for evaluation. 

\section{Method: outcome-driven guidance evolution}
\label{sec:method}

\begin{figure}[t]
\centering
\includegraphics[width=0.9\linewidth]{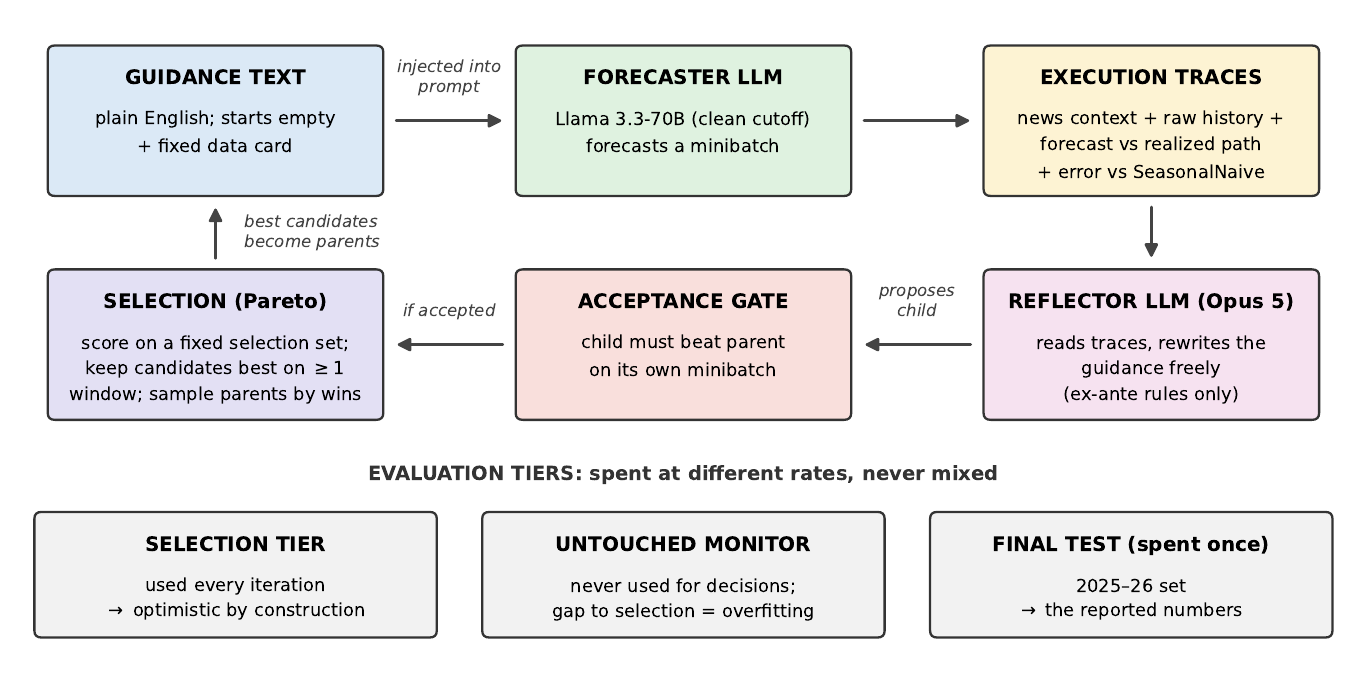}
\caption{The evolution loop: forecast with the current guidance, reflect on traces, gate on a minibatch, select on a fixed set under an instance-level Pareto front; an untouched monitor arbitrates and final tiers are spent once.}
\label{fig:loop}
\end{figure}

The object we optimize is not a model but a \textbf{guidance text} $g$, up to $\sim$1{,}000 words pasted into a fixed prompt template alongside a static, strategy-free \emph{data card} describing the benchmark. One iteration of the loop (Fig.~\ref{fig:loop}):

\begin{enumerate}[leftmargin=14pt, itemsep=1pt, topsep=2pt] \item \textbf{Forecast.} The clean forecaster predicts a minibatch ($B{=}16$) of training windows with $g$ in its prompt, emitting a brief rationale alongside the numbers. \item \textbf{Reflect.} The reflector reads the minibatch's full execution traces (raw histories, complete news context, forecast vs.\ realized paths, per-window MASE, rationales) and rewrites $g$ freely; only ex-ante usability, generality, and the word cap bind. \item \textbf{Gate.} The child must beat its parent on the same minibatch or is discarded. \item \textbf{Select.} Survivors are scored on the full fixed selection set; a candidate stays if best on $\ge$1 window (instance-level Pareto front), parents are sampled by sole-best counts, and every fourth iteration attempts a complementarity-screened \emph{merge}. \end{enumerate}

Nothing is fine-tuned and no gradient exists: learning is the reflector turning observed forecasting mistakes into written advice, and selection keeping only advice that measurably works on windows it was not written from. The artifact is versioned text and the entire ``model update'' is a readable diff.

\section{Main results}
\label{sec:results}

\subsection{The evolved prompt beats TSFMs on held-out data}

\begin{table}[t]
\centering\small
\  \caption{GM MASE on the 249-window final tier (origins Sep-2025 --
  Mar-2026, after every model's cutoff). Same windows, same pipeline,
  same metric for every row. Best per column in bold; $^{*}$marks cells within one standard error of the column best (paired on log-MASE).}
\label{tab:main}
\begin{tabular}{lcccc}
\toprule
arm & Commodity & Currency & SearchTrend & ALL \\
\midrule
Llama 3.3 + \gnum{} (evolved) & \textbf{0.399} & \textbf{0.864}$^{*}$ & 1.422 & \textbf{0.776}$^{*}$ \\
Opus 4.6 + \gtext{} (transferred) & 0.472 & 1.077 & \textbf{1.110} & \textbf{0.785}$^{*}$ \\
Opus 4.6 + \gnum{} (transferred) & 0.528 & 0.905 & 1.161 & \textbf{0.807}$^{*}$ \\
Llama 3.3 + \gtext{} (evolved, text-using) & 0.463 & 0.974 & 1.341 & 0.824 \\
\midrule
TabPFN-TS-3 & 0.654 & \textbf{0.853}$^{*}$ & 1.259 & 0.898 \\
TimesFM-2.5 & 0.711 & 0.911 & 1.256 & 0.939 \\
Chronos-2 & 0.794 & 0.930 & 1.166 & 0.957 \\
\midrule
Llama 3.3 unguided & 0.802 & 1.178 & 1.274 & 1.043 \\
Opus 4.6 unguided & 0.899 & 1.124 & 1.164 & 1.043 \\
SeasonalNaive & 0.816 & 1.244 & 1.437 & 1.115 \\
\bottomrule
\end{tabular}
\end{table}

Table~\ref{tab:main} is the paper's central result: on the most recent, provably uncontaminated data, \textbf{all four evolved cells beat all three zero-shot TSFMs overall} (nominally best: Llama with the $\sim$900-word \gnum{} guidance, GM 0.776 vs.\ TabPFN-TS-3's 0.898). On Currency, the best TSFM is nominally ahead (0.853 vs.\ 0.864), a statistically indistinguishable gap (paired
  $\Delta$ $0.045 \pm 0.039$, evolved arm winning 44\% of windows). \textbf{Both unguided frontier LLMs sit at SeasonalNaive}: raw LLM capability contributes nothing typical-case, and the entire gap between useless and state of the art is carried by a piece of text.

Where does the value come from? It is \emph{domain-concentrated}:
event-heavy daily commodities improve $\sim2\times$ over card-only
baselines, while weekly search barely improves, and on post-refresh
2025--26 SearchTrend windows every method, TSFMs included, is worse
than SeasonalNaive. Reading the evolved text (Appendix~\ref{app:c3})
explains the mechanism: \gnum{} is a regime-first \emph{numeric
policy} (damped random-walk anchors for daily series, calendar-analog
seasonality for weekly ones, explicit spike-decay and off-season
floors), with news demoted to a tightly scoped role: ``dated,
quantified, in-horizon shocks only, applied at half the believed
size.'' TimesX's own benchmark-level insight, that an explicit
no-change option is the single most valuable ingredient, is
rediscovered by evolution and sharpened into quantitative,
situation-scoped caution rules. 

\textbf{Outperforming TimesX benchmark's strongest naive composition.} TimesX's best composed method, the simple TSFM+LLM average, scores 0.979 on our tier, \emph{worse than TimesFM alone} (0.939): post-cutoff, the LLM half contributes SeasonalNaive-grade noise. Averaging a TSFM into the evolved arm also hurts (TabPFN+\gnum: 0.887 vs.\ 0.776); only averaging the two evolved artifacts edges the best single (0.770, within one SE). Composition pays only when every ingredient carries learned guidance (Appendix~\ref{app:avgens}; cf.\ \S\ref{sec:compiled}).

\subsection{Cost: prompt updates versus retraining}

An accepted iteration costs $\approx$\$2 of API calls; a full campaign
with its final evaluation is $\approx$\$50--100 on no dedicated
hardware. The artifact updates by \emph{diff}: adapting to a new
regime means one more reflection over recent traces, not a retraining
run, and every update is a readable, reversible text change. The price
is inference-time length ($\sim$1k extra prompt tokens) and the
pathologies quantified next.

\subsection{Prompt evolution dynamics insights}
\label{sec:learnings}

Three findings replicated across independent experiment designs:

\textbf{(i) Generation-1 dominance.} Essentially all out-of-loop gains arrive in the \emph{first} reflection; 8--13 further operator applications per campaign at best \emph{tied} the champion on the untouched monitor, and merges went 0-for-3 against best-parent baselines. One pass over $\sim$50 windows of traces captures most of what $\sim$900 words can encode here (see \S\ref{sec:discussion} for why this does not obviate the loop).

\textbf{(ii) Selection optimism dominates.} Gains measured on data the
loop consults ran $2$--$3\times$ the untouched-monitor deltas ($\sim$40\%
optimism in the champion experiment). An out-of-loop monitor is a design
necessity for reflective self-improvement.

\textbf{(iii) Error-only objectives suppress the text modality.} Under a MASE-only objective, evolution converges to text-gating guidance and directional recall collapses, rediscovered across experiment designs and frontier executor's own traces. On this corpus, where news is largely redundant with the numeric
  history, the selection is accurate: text slightly hurts guided inference
  (\S\ref{sec:events}) and contributes only calibration during learning
  (\S\ref{sec:reflection-ablation}). Text usage in the guidance survived
  only under an explicit direction objective.

  
  \subsection{Cross-model transfer}
  \label{sec:transfer}
  
  Freezing the guidance and swapping the executor (Table~\ref{tab:main}):
  \gnum{} transferred from Llama to Opus 4.6 scores 0.807 vs.\ 0.776, a
  gap within one standard error (paired on log-MASE), so transfer shows
  no measurable loss on this tier, and both arms remain ahead of every TSFM.



\section{Extended experiment: compiling guidance to code, with a TSFM in
the loop}
\label{sec:compiled}

\begin{table}[t]
\centering\small
\caption{The compiled system vs.\ reference arms: GM MASE on the
249-window final tier (full singles in Table~\ref{tab:main}). ``Naive
avg.'' rows are per-window means of two arms' point forecasts, the
TimesX paper's strongest composition recipe, shown at its best cells
from our 17-pair grid (Appendix~\ref{app:avgens}). Best per column in bold; $^{*}$marks cells within one standard error of the column best (paired on log-MASE).}
\label{tab:compiled}
\begin{tabular}{lcccc}
\toprule
arm & Commodity & Currency & SearchTrend & ALL \\
\midrule
\textbf{\gcode{}} (learned: code + TabPFN + LLM) & 0.608 & \textbf{0.845}$^{*}$ & \textbf{1.042} & \textbf{0.807}$^{*}$ \\
naive avg.\ of Llama+\gnum{} and Llama+\gtext{} & 0.420 & \textbf{0.874}$^{*}$ & 1.316 & \textbf{0.770}$^{*}$ \\
naive avg.\ of TabPFN and Opus+\gtext{} & 0.622 & 0.907 & 1.136 & 0.854 \\
\midrule
best evolved prompt (Llama+\gnum) & \textbf{0.399} & \textbf{0.864}$^{*}$ & 1.422 & \textbf{0.776}$^{*}$ \\
best TSFM (TabPFN-TS-3) & 0.654 & \textbf{0.853}$^{*}$ & 1.259 & 0.898 \\
\bottomrule
\end{tabular}
\end{table}

Reading the evolved champions (Appendix~\ref{app:c3}) reveals that they are heavy on \emph{time-series features} (anchors, damped drifts, seasonal analogs, spike rules) i.e.\ largely \emph{executable} content. This motivates the question: how much of the learned guidance survives being compiled into deterministic code, and can the loop then learn to \emph{use a TSFM as a tool} rather than compete with it?

\textbf{Setup.} The genome is still a guidance text (word cap 6{,}000; it may contain formulas and literal Python) but its phenotype is a two-stage runtime. (1)~A coding LLM (Claude Opus 4.6; all data postdates its cutoff) \emph{compiles} each candidate once into a deterministic \texttt{forecast(window)} function in an offline sandbox against a 12-window dev set; the session is \emph{engineering-only} by contract, capped at 6 tool turns, and the harness independently re-runs the scorer on the final code, so implementation quality can neither block nor fake a candidate. (2)~Per window, the code emits a \emph{proposal}; a final LLM pass sees the window, guidance, code, proposal, and default TabPFN forecast, and makes the final call. The code receives only ex-ante inputs including a precomputed TabPFN-TS-3 forecast and may re-invoke TabPFN live with \emph{engineered inputs} (transformed targets, truncated context, event flags parsed from the text); TabPFN is the only learned model permitted, and the code stage evaluates at zero LLM cost. Internal baselines: \textbf{V0}, raw TabPFN passthrough; \textbf{V0L}, that proposal plus an \emph{unguided} final LLM.


\textbf{What the guidance learned to do with the TSFM.} The champion's text (Appendix~\ref{app:compiled}) partitions windows by route and assigns TabPFN a different role in each: \emph{lead} on smooth daily commodities (weight up to 0.65, rule drift shrunk), \emph{ignore} on staircase series (hold the plateau), \emph{blend} into a strict year-over-year calendar analog on weekly search, and \emph{feed} with engineered inputs where events matter, while the final LLM is held to tight pass-through discipline: an auditable division of labor mirroring the text-gating that pure prompt evolution discovered.

\textbf{Results.} Twelve iterations produced eight accepted candidates. We attribute the longer runway to the richer phenotype: the reflector now debugs \emph{mechanisms} (``the year-over-year ratio was computed against a window centred on the analog, double-counting the seasonal phase; construct it strictly YoY'') rather than only rewriting advice. \gcode{} scores GM 0.807 (Commodity 0.608 / Currency 0.845 / SearchTrend 1.042), ahead of every TSFM, every unguided arm, and every naive
  TSFM-involving ensemble, and statistically tied with the best evolved
  arms overall (within one SE of Llama+\gnum{}, GM ratio 1.039) (Table~\ref{tab:compiled}). Three sub-results stand out: (1) its \textbf{SearchTrend 1.042 is the best of any experiment} (next: 1.110; Llama+\gnum{}: 1.422), the reflector's YoY mechanism being the first real progress; \gcode{} beats every naive TSFM+LLM average (best cell 0.854), so the learned division of labor is worth $\sim$0.05 GM over per-window averaging; and the code stage runs in milliseconds per window, deployable as an auditable program with one optional LLM arbitration.

\section{Ablations}
\label{sec:ablations}

Both ablations run on the final held-out tier, excluding the compiled experiment of \S\ref{sec:compiled}.

\subsection{Forecast horizon: the advantage is front-loaded}
\label{sec:horizon}

We rescore the \emph{same cached 12-step forecasts} at horizons
$h\in\{3,6,12\}$ (the MASE scale is horizon-independent by definition,
so numbers are comparable). No model is re-run.

\begin{table}[h]
\centering\small
\caption{Horizon ablation (GM MASE on the 249-window tier; first $h$
steps of the same cached forecasts). Best per column in bold; $^{*}$marks cells within one standard error of the column best (paired on log-MASE).}
\label{tab:horizon}
\begin{tabular}{lccc}
\toprule
arm & $h{=}3$ & $h{=}6$ & $h{=}12$ \\
\midrule
Llama 3.3 + \gnum{} & \textbf{0.291} & \textbf{0.413} & \textbf{0.776}$^{*}$ \\
Opus 4.6 + \gtext{} & 0.349 & 0.472 & \textbf{0.785}$^{*}$ \\
TabPFN-TS-3 & 0.434 & 0.565 & 0.898 \\
TimesFM-2.5 & 0.453 & 0.594 & 0.939 \\
Llama 3.3 unguided & 0.463 & 0.679 & 1.043 \\
Opus 4.6 unguided & 0.367 & 0.591 & 1.043 \\
\bottomrule
\end{tabular}
\end{table}

The evolved guidance's advantage over the best TSFM \emph{widens} as
the horizon shrinks: 14\% GM at $h{=}12$, 33\% at $h{=}3$ (0.291 vs.\
0.434). This matches the mechanism: damped anchors and level-shift
discipline matter most in the first steps, where TSFM priors mis-set
the level. Unguided Opus 4.6 is competitive at $h{=}3$ (0.367) but
decays fastest: raw LLMs set a reasonable initial level, then invent
trends; the guidance's largest contribution is stopping the ramp.

\subsection{Re-evolving for the short horizon does not beat truncating}
\label{sec:reevolve}

Since the advantage is front-loaded, should one \emph{optimize} for the short horizon rather than truncate? We re-ran the champion campaign's protocol from scratch at $h{=}3$ end-to-end (targets, prompts, MASE) on the same pool (6 iterations, 4 accepted; untouched monitor: $\Delta{=}{-}0.204{\pm}0.045$ vs.\ selection $-0.198$, essentially zero optimism gap). Spending the final tier once at $h{=}3$:

\begin{table}[h]
\centering\small
\caption{Optimize-for-$h{=}3$ vs.\ truncate-from-$h{=}12$ (GM MASE, 249-window tier, $h{=}3$). Best per column in bold; $^{*}$marks cells within one standard error of the column best (paired on log-MASE). In the
  SearchTrend column no cell differs from the best by more than about one standard error:
  the domain does not discriminate methods at this horizon.}
\label{tab:reevolve}
\begin{tabular}{lcccc}
\toprule
arm & Commodity & Currency & SearchTrend & ALL \\
\midrule
\gnum{} ($h{=}12$-evolved, truncated) & \textbf{0.097}$^{*}$ & 0.439 & 0.701 & \textbf{0.291} \\
$h{=}3$-evolved champion & \textbf{0.106}$^{*}$ & 0.464 & 0.739 & 0.311 \\
TabPFN-TS-3 & 0.267 & \textbf{0.405} & 0.727 & 0.434 \\
TimesFM-2.5 & 0.284 & 0.484 & 0.693 & 0.453 \\
unguided & 0.255 & 0.562 & 0.757 & 0.463 \\
\bottomrule
\end{tabular}
\end{table}

Both evolved arms crush the TSFMs at $h{=}3$ on commodities ($\sim$0.10 vs.\ 0.27), but re-evolution does \emph{not} beat truncation: the 12-step champion is better in every domain than the $h{=}3$-evolved one (per column, TSFMs stay competitive at short range, TabPFN taking Currency at 0.405 and TimesFM SearchTrend at 0.693). The $h{=}3$-evolved guidance rediscovers the same core policy (median anchors, damped capped drift, flat regimes, post-spike snap-back): the long-horizon discipline \emph{subsumes} the short-horizon one, and one artifact serves all horizons.

\subsection{Removing text at inference}
  \label{sec:events}
  
  We strip the unstructured text (background and the dated news events)
  from the prompt entirely, keeping structured holiday and sibling fields,
  and re-run the Llama arms on the same tier.
  
  \begin{table}[h]
  \centering\small
  \caption{Events ablation (GM MASE, 249-window tier, Llama arms).}
  \label{tab:events}
  \begin{tabular}{lccc}
  \toprule
  arm & with events & without events & $\Delta$ \\
  \midrule
  unguided (V0) & 1.043 & 1.162 & $+0.119$ \\
  + \gnum{} & 0.776 & 0.761 & $-0.015$ \\
  + \gtext{} & 0.824 & 0.780 & $-0.044$ \\
  \bottomrule
  \end{tabular}
  \end{table}
  
  The result is a clean dissociation: the events are worth $0.119$ GM to
  the \emph{unguided} model (a raw LLM extracts real value from news),
  while for the \emph{evolved} arms the sign flips, removing text slightly
  \emph{helps}. The residual run-time effect of raw text on a guided
  forecaster is net-negative narrative-chasing, and deployment can skip
  context retrieval at zero accuracy cost. One natural reading is that the
  text's value has been \emph{absorbed into the policy} during learning,
  since the reflector reads every trace's full news context when writing
  rules. The next ablation tests this directly.
  
  \subsection{Ablating text from the reflection loop}
  \label{sec:reflection-ablation}
  
  We re-run the campaign in two arms sharing seeded minibatches: arm A's
  reflector sees full execution traces, arm B's sees the same traces with
  all text withheld. The forecaster's prompt is identical in both arms,
  and each frozen champion is spent once on the 249-window tier. Arm A,
  a fresh-seed replica of the original campaign, lands at 0.741 versus
  \gnum{}'s 0.776 (GM ratio $0.954 \pm 0.019$), so seed variance is of the
  same order as the effect and only the paired A--B contrast carries
  causal weight: text-blind B reaches 0.768, worse than A by
  $+0.036 \pm 0.015$ log-MASE (GM ratio 1.037, same sign in all domains),
  yet recovering nearly all of the guided gain (unguided 1.043).     The artifacts explain the gap: B writes news rules from priors alone
  (``Never push to a new record extreme without hard, dated, quantified
  news'') but cannot calibrate them, while A fits amplitudes (``a major
  dated event for that exact topic justifies +30--60\% above baseline in
  the one or two weeks containing it''). Text evidence teaches the
  calibration of event rules. The rest of the policy comes from numeric
  errors and world-knowledge priors.

\section{Continual deployment: a protocol and one executed cycle}
\label{sec:deployment}

\textbf{A continual deployment protocol.} We propose a pre-registered
rolling loop over the deployed guidance $g_t$, run whenever a new
batch of outcomes resolves (period $t$). When period $t$ closes: (1)~\emph{record} $g_t$'s performance on a fresh
monitor sample from $t$; (2)~\emph{re-adapt}: at most four gated
reflection iterations on $t$'s resolved windows; (3)~\emph{promote
conservatively}: ship $g_{t+1}$ only if it beats $g_t$ on held-back
validation; (4)~\emph{deploy}
on period $t{+}1$. The full multi-period run reports deployed MASE
against frozen-$g_1$ and re-evolved-from-scratch baselines, cost, and
forgetting, which is uniquely inspectable here as a text diff of
deleted rules.

\gnum{},
evolved on 2024 data and deployed blind on 2025--26, kept its commodity
and FX gains but broke at the SearchTrend regime shift (1.42,
Table~\ref{tab:main}). We ran steps (2)--(4) on that break: four
reflection iterations over resolved post-refresh search windows promoted the first child, which repaired
the domain-specific performance. Since the update helps the broken domain but slightly harms the others, the natural
  deployment is per-domain: keep the original \gnum{} for commodities and FX and use the
  updated guidance only for SearchTrend.  As a single global artifact the update also exhibits
the stability--plasticity trade-off concretely: small,
text-diff-inspectable forgetting on the daily domains. Routing the update to the broken domain only sets a
new best on the final tier (\textbf{0.760}). Continual re-adaptation
is thus cheap, auditable, and effective, with forgetting managed by
routing.

\section{Discussion and limitations}
\label{sec:discussion}

\textbf{What continual prompt evolution buys:} a weight-free loop with
an auditable, portable, versioned artifact updateable for dollars,
turning a frozen open-weights LLM from SeasonalNaive-equivalent into
the strongest single method here. \textbf{Does generation-1 dominance invalidate the loop?} The plateau tracks \emph{artifact capacity}, not the loop: three protocols sharing the $\le$1k-word text phenotype all plateaued, while the wider phenotype of \S\ref{sec:compiled} kept improving into generation~2, its mutations changing in kind from rephrasing advice to debugging mechanisms. And under generation-1 dominance the loop's role shifts from search to \emph{verification}: selection ran $2$--$3\times$ optimistic and champions were rarely the first child, so the gate and untouched monitor are what make a generation-1 artifact shippable. For continual deployment this is an asset: each quarter is a fresh generation-1 problem, hence $K{\le}4$. Section~\ref{sec:deployment}'s executed cycle grounds the continual claim: a measured regime break was repaired, with forgetting made inspectable and managed by routing; the multi-period rolling run is future work.

\textbf{Limitations and future work.} All current findings rest on one
benchmark family. The caution-first character of
every evolved policy, and the result that guided inference can discard
the news text, may be joint properties of the MASE objective and of
web-retrieved context whose events are often already priced in ~\cite{cik2024}.
Generation-1 sufficiency may likewise reflect the benchmark's
homogeneity (three domains, one window geometry); more heterogeneous
corpora may extend the useful depth of evolution, as the richer code
phenotype already hints. The LLM arms also receive calendar, holiday,
and sibling-series information that pure-sequence TSFMs do not; a
controlled parity study would locate how much of the remaining gap is
information rather than policy. Finally, the rolling deployment of \S\ref{sec:deployment}
carries one executed cycle; the multi-period run is future work.


\bibliographystyle{plain}

\appendix

\newpage
\section{The \gnum{} and \gtext{} evolved prompts}
\label{app:c3}

\gnum{} is the numeric-policy champion and \gtext{} the text-using one. \gnum{} was evolved in a single reflection over a 16-window minibatch of traces, never on the untouched monitor beaten in 10 subsequent iterations. Verbatim original \gnum{}:

\begin{lstlisting}
# FORECAST GUIDANCE

Your dominant historical failure mode is **inventing trends**: taking the last few points' direction (or a news narrative's "bullish/bearish" tone) and extrapolating a smooth 12-step ramp. This lost badly again and again. MASE rewards the **conditional median path**, which for most of these series is close to flat, or close to last year's same-calendar shape. Follow the procedure below.

## 0. Core priors (apply before any reasoning about news)

1. **News text almost never tells you the level path.** Use it only for: (a) scheduled, dated events falling *inside* the horizon; (b) confirming which calendar season you are in. Reports dated weeks/months before the origin are already in the price/index -- do not re-apply them as a directional push.
2. **No monotone 12-step ramps** unless justified by a repeated calendar pattern in the series' own history.
3. Total movement from the last value across the whole horizon should be *small* relative to the series' historical 12-step swings, unless seasonality says otherwise. When in doubt, flatter is better.
4. Never exceed the historical range; respect hard bounds (search indices 0-100; off-season floors such as 1).

## 1. Decide which regime the series is in

**A. Daily series (FX, commodity prices)** -> random-walk regime.
**B. Weekly search indices** -> annual-seasonality / event regime.

## 2. Regime A: daily FX & commodity prices

These are near-random walks with mild mean reversion at 1-3 week horizons. Momentum in the last 5-10 days carries almost no information.

Procedure:
1. Compute `L5 = mean(last 5 obs)`, `L20 = mean(last 20 obs)`, and `last`.
2. **Anchor** = `0.6*last + 0.4*L5`. If `last` deviates from `L20` by more than ~1.5x a typical daily move xsqrt5 (i.e. the last 1-3 points are a sharp jump/drop away from the recent plateau), pull the anchor 30-50% of the way back toward `L20` -- a fresh sharp move is as likely to retrace as to continue.
3. **Drift**: fit a slope over the last ~20 observations, then use **at most 25% of it**, and let it decay to zero by step 6-8. Cap the cumulative 12-step drift at roughly the size of one typical *weekly* move in that series. If the recent window is flat/low-volatility, use zero drift.
4. Output a nearly flat path around the anchor. Repeated identical values in history (stale quotes, discrete price steps, weekends, holidays) mean the series *sticks* -- for such series just repeat the anchor, rounded to the observed price grid, for all 12 steps. Do not step it down or up "to be safe."
5. Only override this with a directional path if the context contains a **dated, quantified, in-horizon** supply/policy shock (e.g. an announced tariff, an official price change effective on a date inside the horizon). Even then, apply half the move you believe in.

Sanity check before submitting: *Would a flat line at the last value beat my path if the series just wobbled?* If your path drifts more than a couple of percent away from the anchor, delete the drift.

## 3. Regime B: weekly search interest

The history is ~96 weeks ~ 1.85 years, so the horizon weeks have an **analog one year earlier at indices (t-52)**. That analog is your single most valuable signal -- far more valuable than the last 3 weeks' direction.

Procedure:
1. Extract the 12 values at positions `t-52 ... t-41` (the same calendar weeks last year) -> `S[1..12]`.
2. Compute a level ratio `r = mean(last 8 obs) / mean(the 8 obs ending at t-52)`. Clip `r` to [0.7, 1.4], then shrink it toward 1: `r' = 1 + 0.6*(r-1)`.
3. Base path = `r' * S[i]`, then **shrink extremes toward the recent local level**: replace each point with `0.7*(r'*S[i]) + 0.3*(recent 4-week mean)`. This keeps the seasonal shape but hedges timing error.
4. Handle spikes explicitly:
   - If `S` contains an isolated one/two-week spike (annual event, championship, sale day) and you can tie it to a **calendar date visible in the context or an obvious annual recurrence**, keep it but at ~60-70% of its scaled amplitude, and put a partial bump (~30%) in the adjacent week.
   - If the spike's timing is uncertain, flatten it to ~35-45% amplitude spread over 2-3 weeks. A misplaced spike costs double under MAE.
   - If last year shows an idiosyncratic one-off spike unlikely to repeat, drop it and use the surrounding baseline.
5. **Post-spike decay rule** (critical): if the recent 5-15 weeks contain an unusually large news-driven spike that is already declining, forecast continued decay toward the **pre-spike baseline** (the median of the 20+ weeks before the spike), reaching it within ~3-5 steps, then flat. Never extrapolate the *rise* of a spike or restart one.
6. **Off-season rule**: if the series is currently at its floor (e.g. near-constant minimum) and the same weeks last year were also at the floor, output the floor value repeated. Do not add wiggle.
7. **Deseasonalized flat series**: if the series has no clear annual pattern and is currently mid-range, output the recent 6-8 week median, essentially flat, with at most a small tilt (<=10% of level) in the direction of the last-year analog's slope.
8. U.S. public holidays inside the horizon: ignore them for weekly search indices and commodity levels unless the topic *is* the holiday. They are not a reason to add fluctuation.

## 4. Final checklist (all domains)

- Is my path a straight ramp? -> Almost certainly wrong; flatten or re-derive from the year-earlier analog.
- Did any part of the news text change my numbers? -> Justify it with a date inside the horizon; otherwise revert.
- Does my first forecast value sit within a typical one-step move of the last observation? -> It must.
- Is the spread of my 12 values plausible versus how much this series has actually moved over 12 steps in its own history? -> If larger, compress toward the anchor.
- Rounding: match the granularity of the history (integers for search indices, the observed price grid for commodities, 4 decimals for FX).

Output exactly the 12 numbers requested; keep reasoning brief and grounded in the numeric history, not the narrative.
\end{lstlisting}

Note what the text is made of: anchors, damping factors, capped drifts, seasonal analogs, spike handling, \emph{time-series features and policies}, not text comprehension. This observation motivated the compiled extension of \S\ref{sec:compiled}.

Verbatim original \gtext{}:

\begin{lstlisting}
**UNITS.** u = one seasonal unit = the MASE denominator's typical step change (estimate: mean absolute change at the seasonal lag over the 96-step history). Every judgement below is stated in u. The call is horizon mean - last value, bucketized at +/-1.0u.

**STEP 0 -- GATES: what to throw away before reasoning.**
1. **Expired subject.** Items whose reported period closed before the origin (last month's print, last season's report, a concluded protest/tour/conference) -> discard for direction.
2. **Wrong-year / scheduled-calendar items.** Notices of future meetings, "committee meets eight times a year", dates outside the horizon -> discard.
3. **Already-absorbed news.** If the series has already made a >=2u move consistent with the item and then flattened, the item is absorbed -> discard. Old news cannot be re-spent as new drift.
4. **Non-mechanical items.** Advocacy, corporate PR, litigation procedure, satisfaction surveys, historical retrospectives, generic macro recaps (CPI/payrolls/PMI/GDP already published) -> discard: these have no reliable sign for the next 12 steps.
5. **Level quotes.** Any absolute price/target/index in text is a different construct than the series; may never set the forecast level.
6. **Mutual cancellation.** Equally fresh and specific items with opposing signs cancel.

**STEP 1 -- REGIME CLASSIFICATION (choose exactly one).**
- **R1 Daily, flat/low-volatility:** last 10 steps within ~1u, no run. Baseline: hold last value.
- **R2 Daily, trending:** >=5 of last 7 steps one direction and last-10 move >=1.5u. Baseline: hold last value plus 50-70% of the 10-step slope, decaying, capped 2.5u total.
- **R3 Weekly index, seasonal:** baseline = mean of last 3, plus 60% of the prior-year analog path (mean of weeks t-51...t-40 minus x_{t-52}) when |analog| >= 1.5u.
- **R4 Post-spike / boundary:** last value in the top or bottom decile of the 96-step range, reached in <=6 steps, or a text-named cause has passed. Baseline: revert 1.0-2.0u toward the trailing 12-step mean.
- **R5 Exhausted move:** last value at the 96-step extreme after a long slide/rally that has since flattened for >=6 steps. Baseline: hold last value; **never** extend the move.

**STEP 1b -- EVENT TAXONOMY (how surviving text modifies the regime baseline).**
- **T1 Physical supply/demand change with a date inside or just before the horizon** (new harvest hitting the market, mine/plant outage, export ban, weather-damaged crop, mandated coverage/tariff effective in-horizon): +/-1.0-1.5u, in the direction of scarcity/glut; timing = start within the first 3 steps.
- **T2 Two or more independent, quantified, <=6-week-old items agreeing in direction:** +/-1.0u. A single item, however dramatic, is +/-0.0-0.5u.
- **T3 Official forecast revisions** (agency/analyst estimates for a season): direction only, +/-0.5u, and only if not contradicted by a fresher revision; never a level anchor.
- **T4 Scheduled in-horizon events for search indices** (holidays, launches, campaigns, sports/seasonal peaks): apply the previous cycle's response at the matching horizon steps, up to +/-2u locally; the whole-horizon mean shift is whatever that shape implies.
- **T5 Monetary/policy divergence text for FX:** by itself +/-0u. It counts only as confirmation of R2, adding +0.5u in the direction the price is already moving.
- **T6 Event-driven attention spikes for search indices:** the news that caused the spike implies decay, not continuation -> supports R4 downward reversion.

**STEP 2 -- DECISION.**
- **D1** Take the regime baseline; add the taxonomy adjustments; cap total deviation of the horizon mean at 2.5u (daily) / 3.0u (weekly).
- **D2 Direction lock:** text may not push against the live numeric run. If T-adjustments oppose R2's slope, keep the numeric side and halve the text side.
- **D3 All-or-nothing:** if the resulting horizon-mean deviation is < 1.3u, publish the flat baseline instead (deviations below the bucket edge only cost error). If >=1.3u, shape the path so the deviation is real: reach ~2/3 of it by step 4 and the full amount by step 8, then hold.
- **D4 Move-call warrants.** A non-flat call is justified when **any** of: R2 fires cleanly; R3's analog is >=1.5u; R4/T6 gap to the trailing mean >=1.5u; T1 fires with the numeric trend not opposing. In these cases you must call the move -- declining to call is a scored error, not caution.
- **D5 Flat-call warrants.** Stay flat when: R1 or R5 holds and no T1/T4 fires; all text was gated; text conflicts (gate 6); or the only argument for a move is a slope smaller than 1.3u over 12 steps.
- **D6 Path hygiene:** no undamped slope extrapolation (worst-performing policy at population level, worst on weekly); no long monotone ramp on a weekly index without an analog behind it; no fine-grained fake volatility.

**FAILURE MODES TO AVOID.**
- **F1 Recall collapse:** treating flat as the safe answer everywhere. D4 exists to force calls; a run with almost no up/down calls is failing half the objective.
- **F2 Stale-narrative drift:** small trends imported from months-old reports, against a flat or opposite recent tape (gates 1-3, D2). This is the dominant way persistence beats us.
- **F3 Level import:** steering toward a figure printed in a report (gate 5).
- **F4 Momentum on the wrong horizon:** applying daily-style trend continuation to weekly indices, where the annual analog dominates (R3, D6).
- **F5 Continuing an exhausted decline into a floor** instead of holding or reverting (R5, R4).
- **F6 Under-committed moves:** deviations of 0.3-1.0u that lose MASE against persistence and earn no directional credit (D3).
- **F7 Macro-report guessing on FX:** deriving direction from generic economic releases with no live price confirmation (gate 4, T5).
\end{lstlisting}

\section{Average-ensemble grid}
\label{app:avgens}

Per-window mean of two methods' point forecasts, rescored on the
249-window tier (GM MASE at $h{=}12$); singles for reference..

\begin{table}[h]
\centering\small
\begin{tabular}{lc}
\toprule
arm & ALL \\
\midrule
AvgEns: \gnum{} + \gtext{} (two evolved) & \textbf{0.770} \\
Llama+\gnum{} (best single) & 0.776 \\
\gcode{} (learned composition, \S\ref{sec:compiled}) & 0.807 \\
AvgEns: TabPFN + Opus+\gtext{} (best TSFM+evolved) & 0.854 \\
AvgEns: TabPFN + \gnum{} & 0.887 \\
TabPFN-TS-3 (best single TSFM) & 0.898 \\
AvgEns: TabPFN + TimesFM (TSFM pair) & 0.909 \\
AvgEns: TimesFM + unguided Llama (the TimesX-paper recipe) & 0.979 \\
\bottomrule
\end{tabular}
\end{table}

\section{Compiled-campaign artifacts}
\label{app:compiled}

The champion guidance \gcode{} opens with a change-log addressed to its own lineage, the reflector debugging \emph{mechanisms} across generations:

\begin{lstlisting}
## 0. What changed and why (read this first)
* **Weekly search trends.** The old "level anchor x damped last-year
  ratio" was built wrongly: the *reference* level for the ratio was a
  median over a window **centred on** the year-ago block, so the ratio
  measured "shape vs. its own annual average" while the level anchor
  already contained the current seasonal phase. The two double-counted
  [...] The fix is a strict **year-over-year (YoY) construction** [...]
  **This is the single biggest improvement; implement it exactly.**
* **Daily smooth commodities.** The 50/50 blend with a slope-
  extrapolating rule lost to plain TabPFN twice [...] **shrink the
  rule's influence** -- TabPFN weight up to 0.65, drift coefficient
  down to 0.20.
* **Staircase dailies.** Flat-at-last is confirmed as the right
  default. [...]
* **Pass-through discipline for the final LLM is confirmed** (all
  measured deviations were zero and the code beat SeasonalNaive on
  6/8). Keep triggers tight.
\end{lstlisting}

It then specifies the code directly, e.g.\ route detection:

\begin{lstlisting}
d61 = diff(x[-61:]); frac_flat = mean(d61 == 0)
nuniq = len(unique(round(x[-61:], 6)))
staircase = (not weekly) and (frac_flat >= 0.30 or nuniq <= 12)
route = "WEEKLY" if weekly else ("STAIR" if staircase else
        ("FX" if domain == "Currency" else "DAILY"))
\end{lstlisting}

and TabPFN's role per route (lead / ignore / blend / feed with
engineered inputs). The compiled function keeps the harness's pre-tested scaffolding
(NaN handling, fallback chain to the precomputed TabPFN path, output
validation) and implements the strategy body; compile sessions under the
engineering-only contract converged in 3--4 tool turns with harness-side
verification.

  \section{Compute resources}
  \label{app:compute}
  No GPUs were self-managed. All LLM inference (forecasting, reflection,
  compilation, arbitration) used a commercial serverless cloud API, interactively for small batches and via batch
  inference for evaluations of $\ge$100 prompts. All local computation
  ran on a single consumer laptop (Apple-silicon CPU, 32\,GB RAM): the
  TSFM baselines (TabPFN-TS-3 at $\sim$1\,s per window on CPU;
  TimesFM-2.5; Chronos-2) and the offline Docker sandbox for compiled
  candidates (4 CPUs, 8\,GB memory per container; a 444-window evaluation
  of a compiled candidate takes 10--15 minutes). Individual runs: one
  evolution iteration costs minutes to $\sim$1 hour
  wall-clock including batch queueing; a full campaign with its final
  evaluation is several hours; the executed re-adaptation
  cycle of \S\ref{sec:deployment} took under two hours.
  Every LLM reply is cached, so reproducing
  any reported number from the released caches requires no API calls.
  
\section{Prompt templates}
\label{app:prompts}

\subsection{Forecaster prompt}
Windows are rendered with the CiK-style template of the benchmark
(system prompt: ``You are a useful forecasting assistant.''); the
evolved guidance is injected into the \texttt{<context>} block.

\begin{lstlisting}
I have a time series forecasting task for you.

Here is some context about the task. Make sure to factor in any
background knowledge, satisfy any constraints, and respect any
scenarios.
<context>
{background, dated news events, sibling statistics, holidays}

{fixed data card}

FORECAST GUIDANCE:
{evolved guidance g}

Before the <forecast> block, write a <rationale> block of 2-4
sentences explaining how you formed the forecast and which parts of
the guidance you applied (or chose not to apply) and why.
</context>

Here is a historical time series in (timestamp, value) format:
<history>
(t1, v1) ... (t96, v96)
</history>

Now please predict the value at the following timestamps: {t97..t108}.
Return the forecast in (timestamp, value) format in between <forecast>
and </forecast> tags.
\end{lstlisting}

\subsection{Reflection prompt (guidance mutation)}
\begin{lstlisting}
You are revising the FORECAST GUIDANCE of a time-series forecasting
assistant on a multimodal benchmark: each window has a 96-step numeric
history and web-retrieved news/event context, across three domains
[...]. Objective: MASE [...].

The forecaster ALWAYS sees the following fixed DATA DESCRIPTION in its
prompt; do NOT spend guidance words repeating anything in it:
<data_description> {fixed data card} </data_description>

CURRENT GUIDANCE ({tag}):
<rubric> {current guidance} </rubric>

EVIDENCE - a minibatch of {n} training windows, evaluated blind with
this guidance. For each: the numeric situation and full history, the
FULL news context available at the origin, the model's 12-step
forecast, the ground truth, and the errors (guidance MASE vs
SeasonalNaive MASE).
{per-window evidence blocks}

Write revised guidance. You have COMPLETE freedom of form and
strategy [...]. The only constraints:
- The guidance must be usable EX-ANTE: it may rely only on what is
  visible at the forecast origin, must contain no specific variable
  names from the evidence, and must not encode memorized post-origin
  outcomes.
- Do not spend words repeating the fixed data description.
- <= 1000 words.
Answer with exactly one revision: <rubric_A>...</rubric_A>
\end{lstlisting}

In the compiled campaign (\S\ref{sec:compiled}) the same skeleton is
extended: the reflector is told its guidance has two consumers (the
compiler and the final LLM), the word cap is 6{,}000, evidence rows
additionally show the code's proposal and the default TabPFN forecast,
and the model policy permits TabPFN (via the wrapper) and classical
statistical fitting but no other learned models.

\subsection{Compile-session system prompt (excerpt,
\S\ref{sec:compiled})}
\begin{lstlisting}
You are an expert Python engineer. Your job: faithfully COMPILE a
forecasting guidance text into a single deterministic Python module.
[workspace layout; forecast(window) contract; input schema]

TURN BUDGET: you have exactly 6 tool turns; every tool result shows
turns_remaining. EXPECTED FLOW (3 turns):
1. Write forecast.py (adapt the template's _strategy()).
2. Run scorer.py.
3. Only if scorer reported ERRORS: fix and rerun. Otherwise skip.

FATAL-DETOUR RULE: the moment scorer.py reports errors=0, your NEXT
message must be the final certification (no tool call). Do NOT
investigate windows whose forecasts are merely inaccurate; accuracy is
the guidance's responsibility, not yours. Inaccuracy is NEVER a
defect; only errors are.

- You are an ENGINEER, not a scientist. Strategy is decided by the
  guidance alone; improving it belongs to the outer loop, not you.
- General logic only: never branch on window["key"] or any specific
  series name; no memorized constants for specific series.
[certification contract: IMPLEMENTATION_VERIFIED dev_mase=<x>]
\end{lstlisting}

\subsection{Final-LLM arbitration block (\S\ref{sec:compiled})}
Appended to the forecaster prompt's context in the two-stage runtime:
\begin{lstlisting}
FORECAST GUIDANCE (also given to the code below):
{evolved guidance}

A deterministic program compiled from this guidance already ran on
this window. Its source:
<code> {compiled forecast.py} </code>
CODE'S PROPOSED FORECAST (12 steps): {proposal}
DEFAULT TABPFN-TS-3 FORECAST:        {tabpfn}

You make the FINAL forecast. Follow the guidance: accept the code's
proposal unless the guidance directs an adjustment for this situation
(you can read the news context with full comprehension; the code
cannot). Unjustified deviations from the proposal have historically
hurt accuracy. Before the <forecast> block, write a <rationale> block
of 2-4 sentences stating whether you kept or adjusted the proposal
and why.
\end{lstlisting}

The re-adaptation cycle of \S\ref{sec:deployment} reuses the
campaign-4 forecaster and reflection templates unchanged.

\end{document}